# Characterizing the Set of Coherent Lower Previsions with a Finite Number of Constraints or Vertices

**Erik Quaeghebeur**

Department of Philosophy
Carnegie Mellon University
Pittsburgh, PA, USA

SYSTeMS Research Group
Ghent University
Gent, Belgium

## Abstract

The standard coherence criterion for lower previsions is expressed using an infinite number of linear constraints. For lower previsions that are essentially defined on some finite set of gambles on a finite possibility space, we present a reformulation of this criterion that only uses a finite number of constraints. Any such lower prevision is coherent if it lies within the convex polytope defined by these constraints. The vertices of this polytope are the extreme coherent lower previsions for the given set of gambles. Our reformulation makes it possible to compute them. We show how this is done and illustrate the procedure and its results.

## 1 INTRODUCTION

Consider a subject facing uncertainty and assume that, to deal with it, she formalizes it as a—possibly abstract—gambling problem. The classical coherence criterion then states that, for the gambles she faces, she should choose her previsions—interpreted as fair prices—so as to avoid a sure loss (de Finetti 1974–1975; Walley 1991, §2.8).[1] The resulting coherent previsions are linear expectation operators.

Now, let us allow our subject to be more cautious and specify possibly non-coinciding lower and upper previsions—interpreted as supremum acceptable buying and infimum acceptable selling prices. A modified coherence condition has to be used to accommodate this (Walley 1991, §2.5). The resulting coherent lower previsions are nonlinear expectation operators. (The so-called conjugacy relationship between lower and upper previsions allows us, in all generality, to only consider one of both.)

Coherent lower previsions are, as models of uncertainty, equivalent to convex sets of coherent previsions, or credal sets. Mathematically, they generalize classical probabilities, possibility distributions, and belief functions, among others. In recent years, there has been an increasing interest from the artificial intelligence community for this general type of model—in both incarnations: credal sets and lower or upper previsions—as regards theory as well as applications (see, e.g., Cozman et al. 2004; De Cooman & Zaffalon 2003; De Cooman et al. 2008; Grünwald & Halpern 2008). This paper's contribution lies mostly in the theoretical realm; results are derived and presented using lower previsions, but some illustrations will make good use of credal sets.

We restrict attention to finite sets of simple gambles, i.e., which take only a finite number of values, so that we can consider them as essentially defined on finite possibility spaces (Walley 1991, §4.2.1). In such a context, the set of linear previsions can be represented by the unit simplex of probability mass functions on the possibility space of elementary events. It is completely characterized by both

(a) a finite number of linear constraints: the probability mass on each elementary event is nonnegative and it sums up to one for the whole possibility space; and

(b) a finite number of extreme linear previsions: those that correspond to the degenerate probability mass functions, which are one on a single elementary event and zero elsewhere—the vertices of the unit simplex.

No similar pair of characterizations was known for coherent lower previsions. In this paper, we present

(a) a finitary formulation of the coherence criterion for lower previsions that—when used as an algorithm—generates a finite set of linear constraints that are sufficient to guarantee coherence (but which may contain constraints that are not strictly necessary); and

(b) a procedure that, starting from these constraints, can be used to obtain the finite set of extreme coherent lower previsions that characterizes the set of all coherent lower previsions on the given finite set of simple gambles.

In general, it is not possible to represent a coherent lower prevision in a simpler way: e.g., both lower and upper probability mass functions and lower and upper probabilities are less expressive uncertainty models (Walley 1991, §2.7).

---

[1] Terminology: a *gamble* is the same thing as a *bet* or a *random variable*; a *sure loss* corresponds to a *Dutch book* or *arbitrage*.

Apart from the linear previsions, there are other special classes of coherent lower or upper previsions for which (one or) both characterizations are known. The most important ones are possibility distributions (Quaeghebeur 2009, §2.2.7), belief functions (Brüning & Dennenberg 2008), and lower probabilities (Quaeghebeur & De Cooman 2008). The ideas of Quaeghebeur & De Cooman (2008) provided us with the foundations for this paper, which generalizes their results from an important subclass to the general case.

The two characterizations are two faces of the fact that the set of all coherent lower previsions—regarded as vectors—is a convex polytope: On the one hand, a convex polytope is the intersection of a set of half-spaces, i.e., those defined by the hyperplanes in which its facets lie. On the other hand, it is the convex hull of its extreme points or vertices. So polytope theory will play an important supporting role in this paper (good references: Grünbaum 1967; Ziegler 1995).

**Overview** We present the material as follows: In Section 2, we reformulate the standard coherence criterion into one that generates a finite sufficient set of constraints and in parallel explain the procedure to obtain extreme coherent lower previsions.[2] In Section 3, we analyze the sets of constraints and extreme coherent lower previsions that we obtain when applying our results to different sets of gambles.[3] We end with some conclusions.

**Notation** The generic notation of the finite possibility space is $\Omega$. An event $A$ is one of its subsets and an elementary event $\omega$ is one of its elements.

A gamble $g$ is a real-valued function on $\Omega$ that maps an elementary event $\omega$ to a payoff $g\omega$. (Parentheses are used for grouping, but not for function application.) The indicator $I_A$ of an event $A$ is the gamble that is one on $A$ and zero elsewhere; $I_\omega := I_{\{\omega\}}$ is the singleton indicator of $\omega$.

A lower prevision $\underline{P}$ is a real-valued functional on a set of gambles $\mathcal{K}$—finite in this paper—that maps each of its elements $g$ to a lower prevision $\underline{P}g$, which can be seen as one of the components of the vector $\underline{P}$ in $\mathbb{R}^{\mathcal{K}}$. The upper prevision $\overline{P}$ conjugate to a lower prevision $\underline{P}$ is defined by $\overline{P}g = -\underline{P}(-g)$. The vacuous lower prevision $\underline{P}_A$ expressing ignorance relative to the event $A$ is defined by $\underline{P}_A g = \min_{\omega \in A} g\omega$. Similarly, for any elementary event $\omega$, the degenerate prevision $P_\omega$ is defined by $P_\omega g = g\omega$.

The set of all linear previsions $\mathcal{P}$ is the convex hull of the set of degenerate previsions: $\mathcal{P} = \text{co}\{P_\omega : \omega \in \Omega\}$. So a linear prevision $P$ is completely defined by its probability mass function $p$ on $\Omega$: $P = \sum_{\omega \in \Omega} p_\omega \cdot P_\omega$, where $p_\omega = PI_\omega$.

We use $\langle \mu, \phi \rangle_{\mathcal{X}}$ as a formal shorthand for bilinear expressions of the type $\sum_{x \in \mathcal{X}} \mu_x \cdot \phi x$. Because of this, the identity function id that maps an object onto itself will be useful.

---

[2]For proofs, see (*Proofs*).

[3]For implementation details and a compilation of resulting output files, see (*Implementation*; *Output*).

# 2 COHERENCE CRITERION REFORMULATION

In this section, we first formulate the standard coherence criterion for lower previsions (§2.1). Next, using a toy example, we introduce some useful concepts from polytope theory and the theory of coherent lower previsions (§2.2). Then, we reformulate the coherence criterion in a stepwise fashion, to end up with a finitary version (§2.3–§2.6).

## 2.1 THE STANDARD COHERENCE CRITERION

A lower prevision is coherent if it avoids sure loss and is internally consistent in the following way: consider a gamble $f$ that dominates a positive linear combination of other gambles, each of which is acceptable to the subject in the sense that her lower prevision for it is nonnegative, then she must accept $f$ as well. A standard way to formalize coherence is (cf. Walley 1991, §2.5, also for justifications):

*Definition 1.* A lower prevision $\underline{P}$ on $\mathcal{K}$ is coherent iff $\langle \lambda, \underline{P} \rangle_{\mathcal{N}} - \underline{P}f \leq \max(\langle \lambda, \text{id} \rangle_{\mathcal{N}} - f)$ for all
- gambles $f$ in $\mathcal{K}$,
- subsets $\mathcal{N}$ of $\mathcal{K}$,
- coefficient vectors $\lambda$ in $(\mathbb{R}_{>0})^{\mathcal{N}}$. ◁

We see that a linear constraint on $\underline{P}$ is generated for each $f$, $\mathcal{N}$ and $\lambda$. To get a finitary coherence criterion, i.e., one that generates only a finite number of constraints, the possible values of $\lambda$ need to be restricted to a finite number.

By playing with the normalization and sign of $\lambda$-components we can obtain an equivalent, shorter formulation:

*Definition 2.* A lower prevision $\underline{P}$ on $\mathcal{K}$ is coherent iff $\langle \lambda, \underline{P} \rangle_{\mathcal{K}} \leq \max \langle \lambda, \text{id} \rangle_{\mathcal{K}}$ for all coefficient vectors $\lambda$ in $\mathbb{R}^{\mathcal{K}}$ with at most one strictly negative component. ◁

## 2.2 CONSTRAINTS, VERTICES & CREDAL SETS

It is useful to have a visual image of what the linear constraints appearing in Definition 2 look like. Consider a toy example, where $\Omega = \{a,b,c\}$ and $\mathcal{K}$ consists of two gambles, $f := (1, 1/2, 0)$ and $g := (0, 2/3, 1)$. Up to normalization, each constraint in $(\underline{P}f, \underline{P}g)$-space is completely determined by its coefficient vector $\lambda = (\lambda_f, \lambda_g)$, so there is only one constraint for each orientation.

We have drawn a finite sufficient set of constraints at the top-left corner of the next page. Even with the max-normalization used, the complete set of constraints would be infinite. Each constraint corresponds to a closed half-space, indicated by stubs drawn on the hyperplane delimiting it. The intersection of all half-spaces is the shaded polytope that corresponds to the set of coherent previsions on $\mathcal{K}$.

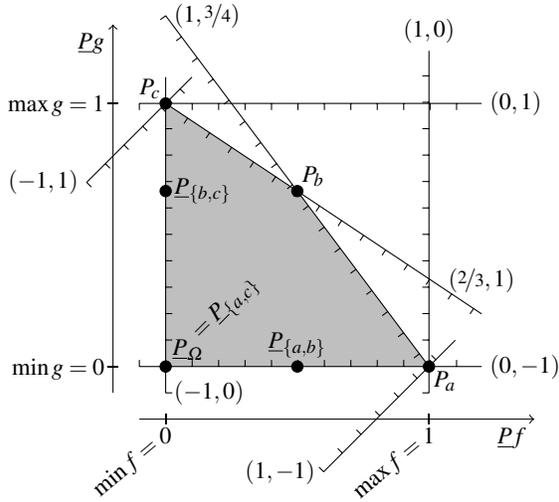

Of the constraints drawn, $(0,1)$, $(1,0)$, $(-1,1)$, and $(1,-1)$ are redundant, meaning that they can be removed without enlarging the intersection. The constraints $(0,-1)$, $(-1,0)$, $(1,3/4)$, and $(2/3,1)$ form a necessary and sufficient set.

Degenerate previsions are coherent lower previsions. So because $\langle \lambda, P_\omega \rangle_\mathcal{K} = \max \langle \lambda, \mathrm{id} \rangle_\mathcal{K}$ for $\omega$ in $\mathrm{argmax} \langle \lambda, \mathrm{id} \rangle_\mathcal{K}$, all generated constraints are strict: the corresponding hyperplanes support the polytope. In our toy example, the only vertex apart from the degenerate previsions is the vacuous prevision $\underline{P}_\Omega$. The other, non-vertex vacuous previsions, $\underline{P}_{\{a,b\}}$ and $\underline{P}_{\{b,c\}}$, have been indicated as well.

The fact that a polytope can be represented both by an H-representation, a set of linear constraints (half-spaces), and a V-representation, a set of vertices (points), is expressed by the so-called Minkowski–Weyl theorem (Ziegler 1995, Thm. 1.1). Going from the H-representation to the V-representation is called vertex enumeration; the other direction is called facet enumeration. Points that can be written as strict convex combinations of vertices and constraints that can be written as positive linear combinations of other constraints are redundant. Using this terminology, our endeavor, schematically, is the following:

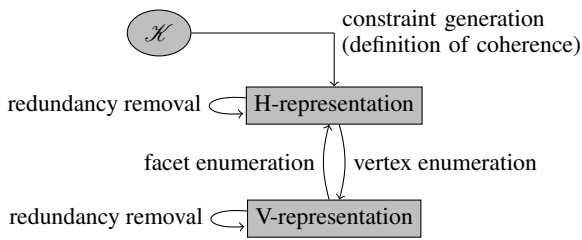

In between the minimal V-representation and the full vector space view, as given above, there is the adjacency graph representation, in which neighboring vertices are connected by edges. It is useful for working with extreme coherent lower previsions. Our toy example's is given on the right.

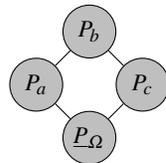

What we have here done by hand, enumerate the vertices of a polytope specified by a set of linear constraints and deduce the adjacency graph, can be done using publicly available polyhedral computation computer programs (e.g., Avis 2000; Fukuda & Prodon 1996).

A standard way to represent a coherent lower prevision $\underline{P}$ is by using its credal set $\mathcal{M}_{\underline{P}}$ (Walley 1991, §3.3). This set consists of all the linear previsions that dominate it: $\mathcal{M}_{\underline{P}} := \{P \in \mathcal{P} : \underline{P} \leq P\}$; and $\underline{P}$ is its lower envelope: for all gambles $g$, $\underline{P}g = \min_{P \in \mathcal{M}_{\underline{P}}} Pg$. The set $\mathcal{P}$ is the unit simplex in $\mathbb{R}^\Omega$ and each one of its points corresponds to a probability mass function. The credal set of a coherent lower prevision $\underline{P}$ is the convex subset of $\mathcal{P}$ determined by the linear constraints corresponding to its values.

For our example, the unit simplex is an equilateral triangle; the constraints are $\langle p, f \rangle_\Omega \geq \underline{P}f$ and $\langle p, g \rangle_\Omega \geq \underline{P}g$. On the side, we give the credal sets of the extreme coherent lower previsions we found.

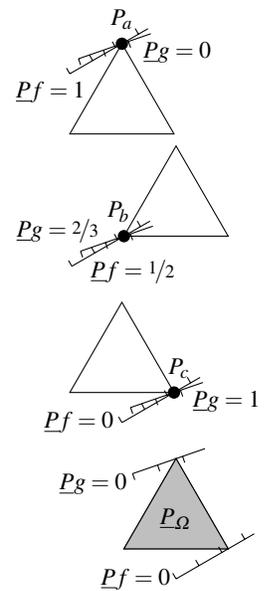

Credal sets represent lower previsions defined on the set of all gambles. This is why in our example, with a finite set $\mathcal{K}$, we had $\underline{P}_\Omega = \underline{P}_{\{a,c\}}$, even though the credal set of $\underline{P}_{\{a,c\}}$ is the convex hull of $P_a$ and $P_c$, a strict subset of $\mathcal{M}_{\underline{P}_\Omega}$. It is prudent to associate coherent lower previsions with the largest credal set possible.[4]

### 2.3 REDUCTION THROUGH NORMALIZATION

A well-chosen normalization of the coefficients is an essential first step in the process towards a finitary coherence criterion. Starting from Definition 2, we normalize the coefficients by normalizing the constraints' right-hand side:

*Definition 3.* A lower prevision $\underline{P}$ on $\mathcal{K}$ is coherent iff $\langle \lambda, \underline{P} \rangle_\mathcal{K} \leq \gamma$ for all
- constants $\gamma$ in $\{-1, 0, 1\}$,
- coefficient vectors $\lambda$ in $\mathbb{R}^\mathcal{K}$
  – with at most one strictly negative component,
  – such that $\max \langle \lambda, \mathrm{id} \rangle_\mathcal{K} = \gamma$. ◁

Now let us return for a moment to the toy example of the previous section. In this example, the constraints $(1,-1)$ and $(1,0)$ are made redundant by the constraints $(0,-1)$ and $(1,3/4)$. Put slightly differently, both are made redundant by $(1, 3/4)$, given $(0, -1)$, i.e., $\underline{P}g \geq \min g = 0$. The right-hand side $\max \langle \lambda, \mathrm{id} \rangle_{\mathscr{K}}$ is one for both $(1, -1)$, $(1, 0)$, and $(1, 3/4)$, so the difference between them, the value of $\lambda_g$, only affects the constraints' left-hand side $\langle \lambda, \underline{P} \rangle_{\mathscr{K}}$. Exactly because $\underline{P}g \geq 0$, we have that $\langle (1, -1), \underline{P} \rangle_{\mathscr{K}} \leq \langle (1, 0), \underline{P} \rangle_{\mathscr{K}} \leq \langle (1, 3/4), \underline{P} \rangle_{\mathscr{K}}$, which explains why $(1, -1)$ and $(1, 0)$ are redundant. The same story can be told about $(-1, 1)$ and $(0, 1)$ versus $(-1, 0)$ and $(2/3, 1)$.

---

[4] This corresponds to using the so-called *natural extension* (Walley 1991, §3.1), the least committal extension to all gambles.

We learn from this that having a set of gambles that only take nonnegative values is useful for removing redundant constraints, by maximally increasing the values of their coefficients in such a way that the right-hand side maximum stays the same, given that we know that the lower prevision must, by coherence, be everywhere nonnegative. Restricting the set of gambles in such a way does not reduce the generality of our endeavor. We next show that we can apply an even more stringent restriction.

Two consequences of coherence are nonnegative homogeneity and constant-additivity (Walley 1991, §2.6.1). Together, for a coherent lower prevision $\underline{P}$ on a set of gambles $\mathscr{K}$, they imply the following: $\underline{P}(\lambda \cdot g + \alpha) = \lambda \cdot \underline{P}g + \alpha$ must hold for all gambles $g$ in $\mathscr{K}$, all nonnegative real $\lambda$ and all real $\alpha$, whenever $\lambda \cdot g + \alpha \in \mathscr{K}$.[5] Equality constraints of this type express that the polytope of coherent lower previsions for a set of gambles in which some gamble pairs are related by such affine transformations is isomorphic to the one for a maximal subset for which this is not the case. So we can restrict our attention to sets of gambles which have a minimum of zero and a maximum of one: For sets of gambles that are not of this type, we can remove the constant gambles and for any other gamble $g$ use $^{g - \min g}/_{\max g - \min g}$ instead.

To modify Definition 3 to take into account what we have learned, we introduce the set of gambles

$$\mathscr{L} := \{ g \in \mathbb{R}^{\Omega} : \min g = 0 \text{ and } \max g = 1 \}.$$

*Definition 4.* A lower prevision $\underline{P}$ on $\mathscr{K}$, a finite subset of $\mathscr{L}$, is coherent iff
  (i) $\underline{P} \geq 0$,
  (ii) $\langle \lambda, \underline{P} \rangle_{\mathscr{K}} \leq \gamma$ for all
     • constants $\gamma$ in $\{0, 1\}$,
     • coefficient vectors $\lambda$ in $\mathbb{R}^{\mathscr{K}}$
        – with at most one strictly negative component,
        – such that $\max \langle \lambda, \mathrm{id} \rangle_{\mathscr{K}} = \gamma$,
        – such that increasing any single component would increase this maximum. ◁

Note that the right-hand side maximum, and thus $\gamma$, cannot be negative. The reason is that all gambles considered are nonnegative and that the single gamble with a strictly negative coefficient—if present—has nonfull support, like any other gamble in $\mathscr{K}$.

## 2.4 ADDING SINGLETON INDICATORS

While Definition 4 does provide a reduction in the number of constraints, we are still faced with an infinity of them. In our toy example, all convex combinations of $(1, 3/4)$ and $(2/3, 1)$ satisfy the definition—they correspond to the constraints through $P_b$. Moreover, the last two restrictions on the coefficient vectors are bothersome to check in practice.

We are going to improve this impractical formulation here. If we add all singleton indicators to the set of gambles $\mathscr{K}$, the requirement that increasing any one component of $\lambda$ must engender an increase of $\max \langle \lambda, \mathrm{id} \rangle_{\mathscr{K}}$, leads to a selection of coefficients that makes $\langle \lambda, \mathrm{id} \rangle_{\mathscr{K}}$ constant (cf. Walley 1991, §A2): Consider a vector of coefficients for which this is not the case; then we can modify it by increasing the coefficients of the singleton indicators until it is. This leads to:

*Definition 5.* A lower prevision $\underline{P}$ on $\mathscr{K}$, a finite subset of $\mathscr{L}$ that contains all singleton indicators, is coherent iff
  (i) $\underline{P} \geq 0$,
  (ii) $\langle \lambda, \underline{P} \rangle_{\mathscr{K}} \leq \gamma$ for all
     • constants $\gamma$ in $\{0, 1\}$,
     • coefficient vectors $\lambda$ in $\mathbb{R}^{\mathscr{K}}$
        – with at most one strictly negative component,
        – such that $\langle \lambda, \mathrm{id} \rangle_{\mathscr{K}} = \gamma$.[6] ◁

So now, to find the $\lambda$ that satisfy the given restrictions, we must find the solution set of the linear system $\langle \lambda, \mathrm{id} \rangle_{\mathscr{K}} = \gamma$ and retain those solutions $\lambda$ that have at most one negative component. However, it is still not practical to implement, and the resulting number of constraints is still infinite. So we need further modifications, to be applied in upcoming subsections. In the rest of this subsection, we show that we have not lost any generality by adding singleton indicators.

To see its effect, let us look how it impacts our toy example:[7] We study the adjacency graph and the $(\underline{P}f, \underline{P}g)$-plane, to which orthogonal projections of $P_a$, $P_b$, and $P_c$ are added.

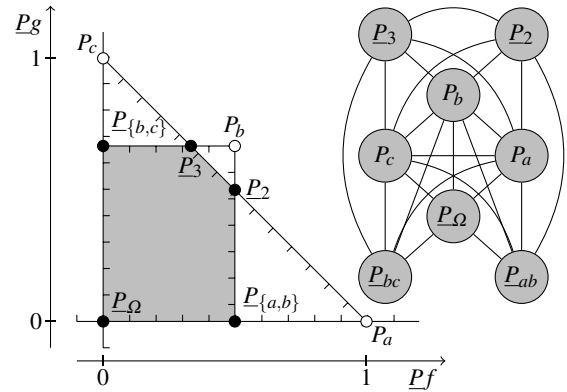

---

[5]Put differently, $\lambda \cdot \underline{P}g + \alpha$ is the unique coherent extension of $\underline{P}$ to $\lambda \cdot g + \alpha$ (gamble-constant addition evaluated pointwise).

[6]The gamble-constant equality is evaluated pointwise.

[7]The data for examples such as this (*Output*, toy) were gathered using our implementation of our finitary coherence criterion (*Implementation*) and publicly available polyhedral computation computer programs (e.g., Avis 2000; Fukuda & Prodon 1996).

The degenerate previsions are the only extreme coherent lower previsions that do not lie in the $(\underline{P}f, \underline{P}g)$-plane; each one of them is moreover connected to all other extreme coherent lower previsions (cf. graph). These observations about the degenerate previsions are general in character:

*Proposition 1.* Of the extreme coherent lower previsions on a set of gambles $\mathscr{K}$ that includes the indicator $I_\omega$ of some elementary event $\omega$ of $\Omega$, the degenerate prevision $\underline{P}_\omega$ is the only one that is nonzero in $I_\omega$. ◁

Four new extreme points have appeared: $\underline{P}_2$, $\underline{P}_{\{a,b\}} = \underline{P}_{ab}$, $\underline{P}_{\{b,c\}} = \underline{P}_{bc}$, and $\underline{P}_3$. The credal sets are given on the right.

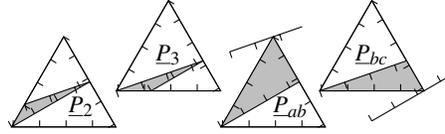

When orthogonally projecting the whole polytope onto the $(\underline{P}f, \underline{P}g)$-plane, the situation pictured in Subsection 2.2 results and these four points become redundant. In general, we can look at the coherent set of lower previsions on any set of gambles $\mathscr{K}$ as the projection on $\mathbb{R}^{\mathscr{K}}$ of the polytope for the set of all gambles. After projection, some vertices become redundant, but the projected polytope is the same as the one that would be obtained by constructing it directly.

It is possible to do the projection in terms of the linear constraints using Fourier–Motzkin elimination (Ziegler 1995, Thm. 1.4). This algorithm is also implemented in publicly available polyhedral computation computer programs (e.g., Avis 2000; Fukuda & Prodon 1996). We now know that no generality is lost by adding the singleton indicators; it only result in an extra processing step. As a consequence, our endeavor schematic must be expanded:

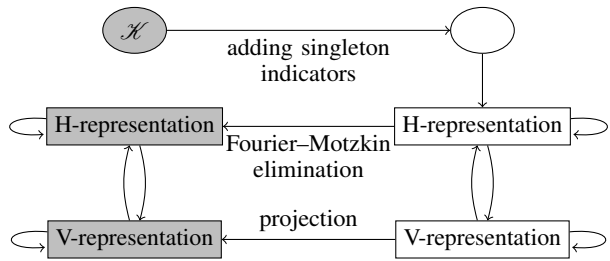

### 2.5 ELIMINATING LINEAR DEPENDENCE

Definition 5 is still not finitary. To apply the essential step, we reformulate it back into a flavor that is closer to Definition 1, in which all coefficients are nonzero:

*Definition 6.* A lower prevision $\underline{P}$ on $\mathscr{K}$, a finite subset of $\mathscr{L}$ that contains all singleton indicators, is coherent iff
  (i) $\underline{P} \geq 0$,
  (ii) $\langle \lambda, \underline{P} \rangle_{\mathscr{N}} \leq \gamma$ for all
    • constants $\gamma$ in $\{0,1\}$,
    • subsets $\mathscr{N}$ of $\mathscr{K}$,
    • coefficient vectors $\lambda$ in $(\mathbb{R}_{\neq 0})^{\mathscr{N}}$
      – with at most one strictly negative component,
      – such that $\langle \lambda, \mathrm{id} \rangle_{\mathscr{N}} = \gamma$. ◁

The following technical lemma tells us that we only need to consider specific subsets of gambles.

*Lemma 1 (Elimination of linear dependence).*[8]
Consider a constraint $\langle \lambda, \underline{P} \rangle_{\mathscr{N}} \leq \gamma$ on $\underline{P}$ determined by
• a set of gambles $\mathscr{N}$ s.t. $(\mathrm{id} - \underline{P})\mathscr{N}$ is linearly dependent,[9]
• a coefficient vector $\lambda$ in $(\mathbb{R}_{\neq 0})^{\mathscr{N}}$,
• a real number $\gamma$ such that $\langle \lambda, \mathrm{id} \rangle_{\mathscr{N}} = \gamma$.
It is equivalent to a constraint $\langle \check{\lambda}, \underline{P} \rangle_{\check{\mathscr{N}}} \leq \check{\gamma}$ determined by
• a set of gambles $\check{\mathscr{N}} \subset \mathscr{N}$ such that $(\mathrm{id} - \underline{P})\check{\mathscr{N}}$ is linearly independent,
• a coefficient vector $\check{\lambda}$ in $(\mathbb{R}_{\neq 0})^{\check{\mathscr{N}}}$ such that $\lambda_g$ and $\check{\lambda}_g$ have the same sign for all $g$ in $\check{\mathscr{N}}$,
• a real number $\check{\gamma}$ such that $\langle \check{\lambda}, \mathrm{id} \rangle_{\check{\mathscr{N}}} = \check{\gamma} \in \{-1, 0, 1\}$. ◁

So we only need to consider the subsets $\mathscr{N}$—finite in number—that satisfy a linear independence condition:

*Definition 7.* A lower prevision $\underline{P}$ on $\mathscr{K}$, a finite subset of $\mathscr{L}$ that contains all singleton indicators, is coherent iff
  (i) $\underline{P} \geq 0$,
  (ii) $\langle \lambda, \underline{P} \rangle_{\mathscr{N}} \leq \gamma$ for all
    • constants $\gamma$ in $\{0,1\}$,
    • subsets $\mathscr{N}$ of $\mathscr{K}$ such that $(\mathrm{id} - \underline{P})\mathscr{N}$ is linearly independent,
    • coefficient vectors $\lambda$ in $(\mathbb{R}_{\neq 0})^{\mathscr{N}}$
      – with at most one strictly negative component,
      – such that $\langle \lambda, \mathrm{id} \rangle_{\mathscr{N}} = \gamma$. ◁

This finitary definition is useful as such for checking the coherence of a known lower prevision $\underline{P}$,[10] but it does not provide an explicit characterization: the selection of subsets $\mathscr{N}$ depends on a condition in which $\underline{P}$ appears, so it cannot be used to generate the constraints that delimit the set of all coherent lower previsions. We remedy this next.

### 2.6 REQUIRING LINEAR INDEPENDENCE

In Definition 7, we make no substantive distinction between the cases for the different values for $\gamma$. But in fact, these determine the nature of the linear system of equations $\langle \lambda, \mathrm{id} \rangle_{\mathscr{N}} = \gamma$ that defines the coefficient vector $\lambda$ for each subset of gambles $\mathscr{N}$. For $\gamma = 1$, this system is inhomogeneous; for $\gamma = 0$, it is homogeneous.

Due to solution unicity considerations, we transform the homogeneous systems to inhomogeneous ones: We bring the single gamble with a negative coefficient—which by construction is always present in this case—to the system's right-hand side and renormalize the coefficients.

---

[8] This lemma is inspired by Walley (1991, §A1).
[9] Notation: $(\mathrm{id} - \underline{P})\mathscr{N}$ is the image of $\mathscr{N}$ under $\mathrm{id} - \underline{P}$.
[10] It is very similar to a criterion by Walley (1991, §A2). Alternatives using linear programming algorithms also exist (see, e.g., Walley et al. 2004, which also deals with contingent gambles).

*Definition 8.* A lower prevision $\underline{P}$ on $\mathcal{K}$, a finite subset of $\mathcal{L}$ that contains all singleton indicators, is coherent iff
(i) $\underline{P} \geq 0$,
(ii) $\langle \lambda, \underline{P} \rangle_{\mathcal{N}} \leq \underline{P}f$ for all
  - subsets $\mathcal{N}$ of $\mathcal{K}$,
  - gambles $f$ in $\mathcal{K} \setminus \mathcal{N}$,
  - coeff. vectors $\lambda$ in $(\mathbb{R}_{>0})^{\mathcal{N}}$ such that $\langle \lambda, \mathrm{id} \rangle_{\mathcal{N}} = f$.
(iii) $\langle \lambda, \underline{P} \rangle_{\mathcal{N}} \leq 1$ for all
  - subsets $\mathcal{N}$ of $\mathcal{K}$ such that $(\mathrm{id} - \underline{P})\mathcal{N}$ is linearly independent
  - coefficient vectors $\lambda$ in $(\mathbb{R}_{\neq 0})^{\mathcal{N}}$
    – with at most one strictly negative component,
    – such that $\langle \lambda, \mathrm{id} \rangle_{\mathcal{N}} = 1$. ◁

Note that we have dropped the restriction to linearly independent $(\mathrm{id} - \underline{P})\mathcal{N}$ in (ii); it is of no more use to us there.

However, the linear independence criterion in (iii) is of use: the following technical lemma allows it to be reformulated to not include reference to the lower prevision $\underline{P}$ anymore.

*Lemma 2 (Preservation of linear independence).* Consider a set of gambles $\mathcal{N}$, a coefficient vector $\lambda$ in $\mathbb{R}^{\mathcal{N}}$ such that $\langle \lambda, \mathrm{id} \rangle_{\mathcal{N}} = 1$, and a lower prevision $\underline{P}$ on $\mathcal{N}$. If the set $(\mathrm{id} - \underline{P})\mathcal{N}$ is linearly independent, then so is $\mathcal{N}$. ◁

The restriction to linearly independent $\mathcal{N}$ can be applied to (ii) as well. The reason is the following: If $\langle \lambda, \mathrm{id} \rangle_{\mathcal{N}} = f$, the nonnegative gamble $f$ lies in the convex conical hull of the set of nonnegative gambles $\mathcal{N}$. Carathéodory's theorem (Ziegler 1995, Thm. 1.15) then tells us that $f$ lies in the convex conical hull of some linearly independent subset of $\mathcal{N}$.[11] Any solution for $\mathcal{N}$ will then be a convex combination of the solutions for the minimal such subsets. The corresponding constraint is the same convex combination of the constraints corresponding to these minimal subsets, and thus redundant.

At last we arrive at our final, finitary and practical definition:

*Definition 9.* A lower prevision $\underline{P}$ on $\mathcal{K}$, a finite subset of $\mathcal{L}$ that contains all singleton indicators, is coherent iff
(i) $\underline{P} \geq 0$,
(ii) $\langle \lambda, \underline{P} \rangle_{\mathcal{N}} \leq \underline{P}f$ for all
  - linearly independent subsets $\mathcal{N}$ of $\mathcal{K}$ such that $1 < |\mathcal{N}| \leq |\Omega|$,
  - gambles $f$ in $\mathcal{K} \setminus \mathcal{N}$ such that $\mathrm{supp}\, f = \mathrm{supp}\, \mathcal{N}$,[12]
  - coeff. vectors $\lambda$ in $(\mathbb{R}_{>0})^{\mathcal{N}}$ s.t. $\langle \lambda, \mathrm{id} \rangle_{\mathcal{N}} = f$.
(iii) $\langle \lambda, \underline{P} \rangle_{\mathcal{N}} \leq 1$ for all
  - linearly independent subsets $\mathcal{N}$ of $\mathcal{K}$ such that $1 < |\mathcal{N}| \leq |\Omega|$ and $\mathrm{supp}\, \mathcal{N} = \Omega$.

---

[11] Carathéodory's theorem can be illustrated by using a section of the cone and the rays corresponding to the gambles considered:

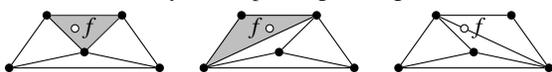

In this example there are two three-dimensional and one two-dimensional minimal subsets of linearly independent gambles.

[12] Notation: $\mathrm{supp}\, \mathcal{N} = \bigcup_{g \in \mathcal{N}} \mathrm{supp}\, g$ and $\mathrm{supp}\, f$ is $f$'s support.

- coefficient vectors $\lambda$ in $(\mathbb{R}_{\neq 0})^{\mathcal{N}}$
  – with at most one strictly negative component,
  – such that $\langle \lambda, \mathrm{id} \rangle_{\mathcal{N}} = 1$. ◁

We have made some cardinality and support restrictions explicit in this definition as well. The support restrictions follow from the restricted choice of coefficients and the gamble equation. The lower bound on the cardinality of the $\mathcal{N}$ comes from the fact those same equalities cannot be satisfied for $|\mathcal{N}| = 1$, as $f \notin \mathcal{N}$ and $I_\Omega \notin \mathcal{L}$. The upper bound is a consequence of the linear independence of the $\mathcal{N}$.

Definition 9 allows us to obtain a characterization of the set of coherent lower previsions in terms of a finite number of constraints and—via vertex enumeration—extreme coherent lower previsions. It can also be used to efficiently check the coherence of a large number of lower previsions on the same set of gambles, as the constraints do not depend on the lower prevision to be checked.[13]

## 3 CONSTRAINTS & VERTICES ANALYZED

In this section, we present a number of cases—sets of gambles $\mathcal{K}$—for which we use Definition 9 to generate a sufficient set of coherence constraints and calculate the corresponding extreme coherent lower previsions. In general, Definition 9 still generates redundant sets of constraints. Applying a redundancy removal algorithm is therefore advisable to speed up the Fourier–Motzkin elimination and vertex enumeration steps.

As a warm-up exercise, and to illustrate some basic ideas, we first look at cases with few—one, two, or three—gambles (§3.1). Then we increase the number of gambles and move the focus to combinatorial aspects (§3.2 and §3.3).

### 3.1 SMALL SETS OF GAMBLES

**One gamble** For completeness, we first consider the almost trivial case in which $\mathcal{K}$ consists of just one gamble, e.g., $f := (1, {}^{1}/{2}, 0)$ on $\Omega := \{a, b, c\}$ (*Output*, 1on3). The one-dimensional nature of of this case inevitably leads to two irredundant constraints and two vertices. From how the credal sets of the example—given on the right—arise, we can deduce that in general the vertices are the vacuous prevision and a vacuous prevision relative to some event.

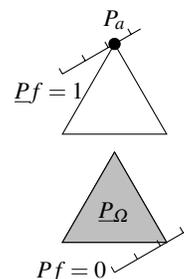

---

[13] It is a mathematical generalization and historical descendant of the one for lower probabilities by Walley (1991, §A3).

**Two gambles**  The toy example of Subsection 2.2 belongs to this class. We found four irredundant constraints and vertices. These numbers reappeared for the set of gambles $\mathscr{K}$ consisting of $f := (1, 1/2, 0)$ and $g := (0, 1, 1/2)$ on $\Omega := \{a, b, c\}$ (*Output*, 2on3) and also when increasing the size of the possibility space—to four, five (*Output*, 2on4, 2on5). A general result for gambles 'in general position'?

As a showcase, we here consider a situation in which both an upper and a lower prevision for a single gamble are specified. This is done by choosing an essentially negation invariant set of gambles, because of conjugacy: $\overline{P}f = -\underline{P}(-f) = 1 - \underline{P}(1-f)$. So we choose the set $\mathscr{K} := \{f, 1-f\}$, with the gamble $f$ as above (*Output*, 1on3_lu). We find three irredundant constraints and vertices. On the right we give their credal sets.

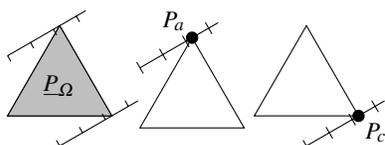

**Three gambles**  Consider a set of gambles $\mathscr{K}$ consisting of $f := (1, 0, 1/2)$, $g := (0, 1/2, 1)$, and $h := (1/2, 1, 0)$ (*Output*, 3on3). There are seven irredundant constraints. We get the following seven-verticed three-dimensional polytope:

|  | $f$ | $g$ | $h$ |
|---|---|---|---|
| $\underline{P}_\Omega$ | 0 | 0 | 0 |
| $\underline{P}_{ab}$ | 0 | 0 | 1/2 |
| $\underline{P}_{ac}$ | 1/2 | 0 | 0 |
| $\underline{P}_{bc}$ | 0 | 1/2 | 0 |
| $\underline{P}_a$ | 1 | 0 | 1/2 |
| $\underline{P}_b$ | 0 | 1/2 | 1 |
| $\underline{P}_c$ | 1/2 | 1 | 0 |

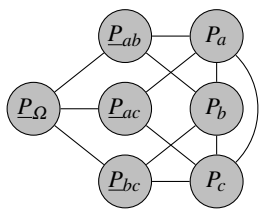

When increasing the size of the possibility space—to four, five—, we observe that the number of irredundant constraints and vertices does not increase substantially (i.e., from 7 to 9) (*Output*, 3on4, 3on5). Up until now, we observed that the number of irredundant constraints and vertices increases superlinearly with $|\mathscr{K}|$.

### 3.2 CLASSES OF EVENT-BASED GAMBLES

To get a feel for larger sets of gambles, we investigate the combinatorics of some classes of event-based gambles, which give rise to interesting classes of lower previsions.

**Singletons**  When $\mathscr{K}$ consists of all $|\Omega|$ singleton indicators, the resulting class of lower previsions is defined by a lower probability mass function (*Output*, l). There are $|\Omega|+1$ irredundant constraints and vertices, which are the degenerate previsions and the vacuous lower prevision.

**Singleton complements**  When $\mathscr{K}$ consists of all $|\Omega|$ singleton complement indicators, the resulting class of lower previsions is—through conjugacy—defined by an upper probability mass function (*Output*, u). There are $2 \cdot |\Omega|+1$ irredundant constraints (but $|\Omega|$ for $|\Omega| = 2$) and $2^{|\Omega|}-1$ vertices, the vacuous lower previsions relative to all events.

This class does not encompass the previous one, even though this is the case for their sets of vertices.[14] This is illustrated on the right by giving the credal sets of the same convex combination of vertices for the former, respectively the latter class.

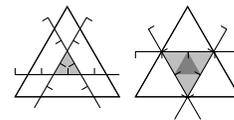

**Singletons and their complements**  When $\mathscr{K}$ consists of all $2 \cdot |\Omega|$ singleton and singleton complement indicators (but $|\Omega|$ for $|\Omega| = 2$), the resulting class of lower previsions is defined by lower and upper probability mass functions (*Output*, lu). We have gathered the number of irredundant constraints (#$\lambda$) and vertices (#$\underline{P}$) for cardinalities two to ten in the table below.

| $|\Omega|$ | 2 | 3 | 4 | 5 | 6 | 7 | 8 | 9 | 10 |
|---|---|---|---|---|---|---|---|---|---|
| #$\lambda$ | 3 | 9 | 16 | 20 | 24 | 28 | 32 | 36 | 40 |
| #$\underline{P}$ | 3 | 8 | 20 | 47 | 105 | 226 | 474 | 977 | 1991 |

After an initial transient, the irredundant constraints fall into a clear pattern: #$\lambda = 4 \cdot |\Omega|$. There is a combinatorial explosion—with no pattern apparent—in the vertex counts.

**Power set**  When $\mathscr{K}$ consists of the $2^{|\Omega|}-2$ indicators for all nontrivial events, the resulting class of lower previsions is defined by an imprecise probability (*Output*, pset). We have now gathered the number of irredundant constraints and vertices for cardinalities two to five in a table.[15] The total number of constraints generated by Definition 9 is given in parentheses.

| $|\Omega|$ | 2 | 3 | 4 | 5 |
|---|---|---|---|---|
| #$\lambda$ | 3 (3) | 9 (17) | 48 (179) | 285 (7351) |
| #$\underline{P}$ | 3 | 8 | 402 | > 1743093 |

Here, there is a combinatorial explosion in the number of irredundant constraints and, quite pronounced, in the number of vertices and generated constraints.

### 3.3 VALUES-BASED GAMBLES

To close off our analysis, we vary the number of gambles for the case $|\Omega| = 3$.[15] The sets of gambles $\mathscr{K}$ we consider to this end contain those gambles in $\mathscr{L}$ that take values in $\{\ell/k : 0 \leq \ell \leq k\}$, for $k$ from one to six (*Output*, vb).[16] We find (for $k$ equal to one, see above):

| $k$ | 2 | 3 | 4 | 5 | 6 |
|---|---|---|---|---|---|
| $|\mathscr{K}|$ | 12 | 18 | 24 | 30 | 36 |
| #$\lambda$ | 15 (178) | 21 (699) | 27 (1796) | 33 (3685) | 39 (6582) |
| #$\underline{P}$ | 49 | 180 | 455 | 928 | 1653 |

---

[14]This is due to the fact that in general pointwise natural extension of the convex set of coherent lower previsions on a given set of gambles to a larger one does not preserve its convexity.

[15]Calculations for higher cardinalities is too computationally taxing for current PCs, i.e., takes more than a few days on a 2006 Intel T7200 processor, or requires more than 2GiB of RAM.

[16]Our tests showed that the specific, structured choice of $\mathscr{K}$ due to the use of evenly spaced values is mostly irrelevant to the combinatorics (*Output*, vb3_2+3, vb3_2like).

The patterns we observe are that $\#\lambda = 3 \cdot (2 \cdot k + 1)$, so the number of irredundant constraints remains remarkably low in this case, and that $\#\underline{P} = (3 \cdot k + 1) \cdot (3 \cdot k^2 - 4 \cdot k + 3)$, the vertex count increases cubicly in $k$.

## 4 CONCLUSIONS

In this paper, we have derived a finitary coherence criterion (Definition 9) for coherent lower previsions defined on a finite set of gambles on a finite possibility space. In this definition, this set is restricted to be a subset of a specific normed class and to contain singleton indicators. However, we have also shown that neither restriction is substantive: the first because extension to the whole class is an isomorphism, the second because Fourier–Motzkin elimination can be applied. In this sense we have given a most general finitary characterization of the set of coherent lower previsions.

From the combinatorial data we obtained, we see that the set of constraints generated by this criterion is still redundant, but that the irredundant sets can be relatively small. The less redundant, the more efficient and thus useful this type of criterion becomes for checking coherence. So working towards an irredundant criterion could be an interesting research path. The regularity found in the values-based gambles case provides hope that progress can be made here.

In this paper, we also showed how the criterion, together with vertex enumeration algorithms, can be used to compute extreme coherent lower previsions. Again, this is a most general procedure in the sense that we can in principle now calculate them for all finite sets of simple gambles and so obtain an alternative characterization of the corresponding set of coherent lower previsions. However, the computational burden—both in calculation time and vertex storage—seems prohibitive for real-life problems.

Applications of extreme coherent lower previsions—e.g., approximation—would rely on our ability to decompose any coherent lower prevision into a convex combination of extreme ones. This decomposition itself is also computationally intensive, further depressing application hopes. Looking at things from the polytope theory side, this decomposition is based on the Minkowski sum of polytopes (Grünbaum 1967, Ch. 15). There is, however, also the Blaschke sum of polytopes, which provides a decomposition in terms of simplices (Alexandrov et al. 2005; Grünbaum 1967, Thm. 15.3.1). It could provide a more fruitful path.

For lower previsions that have to satisfy some additional property, e.g., permutation invariance, the extreme coherent lower previsions can be obtained by adding the necessary constraints to the H-representation. Furthermore, the ideas applied in this paper can be used to derive a similar finitary criterion for the more general conditional lower previsions, i.e., when contingent gambles are considered as well.

## Acknowledgements

This research is supported by a Francqui Fellowship of the Belgian American Educational Foundation. I wish to thank Enrique Miranda and the reviewers for useful comments and suggestions, and Carnegie Mellon for its hospitality.